\pdfoutput=1

\documentclass[11pt]{article}

\usepackage[final]{acl}

\usepackage{times}
\usepackage{latexsym}

\usepackage[T1]{fontenc}

\usepackage{microtype}

\usepackage{inconsolata}

\usepackage{graphicx}

\usepackage{multirow}

\usepackage{latexsym}
\usepackage{amssymb}
\usepackage{amsmath}
\usepackage{amsthm}
\usepackage{booktabs}
\usepackage{enumitem}
\usepackage{graphicx}
\usepackage{color}
\usepackage{comment}
\usepackage{times}
\usepackage{soul}
\usepackage{url}
\usepackage{hyperref}
\usepackage[utf8]{inputenc}
\usepackage{caption}
\usepackage{algorithm}
\usepackage{algorithmic}
\usepackage[switch]{lineno}

\usepackage{subfig}
\usepackage{amsfonts}
\usepackage{xspace}
\usepackage{caption}
\usepackage{diagbox}
\usepackage{makecell}

\newcommand{\debert}[0]{\textsc{Deberta-v3-large}\xspace}

\newcommand{\dolly}[0]{\textsc{Dolly}\xspace}
\newcommand{\msmarco}[0]{\textsc{MS-MaRCo}\xspace}
\newcommand{\gooaq}[0]{\textsc{GooAQ}\xspace}
\newcommand{\narrqa}[0]{\textsc{NarrativeQA}\xspace}
\newcommand{\tweetqa}[0]{\textsc{TweetQA}\xspace}

\newcommand{\default}[0]{\textsc{Default}\xspace}
\newcommand{\limitcat}[0]{\textsc{Limit-Cat}\xspace}
\newcommand{\limit}[0]{\textsc{Limit-Len}\xspace}
\newcommand{\brief}[0]{\textsc{BRIEF}\xspace}

\newcommand{\predreslen}[0]{\textsc{PredResLen}\xspace}
\newcommand{\goldreslen}[0]{\textsc{GoldResLen}\xspace}
\newcommand{\incontxt}[0]{\textsc{BM25-InContext}\xspace}
\newcommand{\similarbm}[0]{\textsc{BM25-length}\xspace}

\newcommand{\minans}[0]{\textsc{MinAns}\xspace}
\newcommand{\addinfo}[0]{\textsc{AddInfo}\xspace}
\newcommand{\explain}[0]{\textsc{Explain}\xspace}
\newcommand{\converse}[0]{\textsc{Convers}\xspace}
\newcommand{\redinfo}[0]{\textsc{RedInfo}\xspace}

\newcommand{\irrel}[0]{\textsc{Irrel}\xspace}

\newcommand{\mans}[0]{\textsc{MinAns}\xspace}

\newcommand{\mansred}[0]{\textsc{MAddNoRed}\xspace}

\title{Brevity is the soul of sustainability: Characterizing LLM response lengths}

\author{
 \textbf{Soham Poddar\textsuperscript{1}*},
 \textbf{Paramita Koley\textsuperscript{2}*},
 \textbf{Janardan Misra\textsuperscript{3}},
 \textbf{Sanjay Podder\textsuperscript{3}},\\
 \textbf{Navveen Balani\textsuperscript{3}},
 \textbf{Niloy Ganguly\textsuperscript{1}},
 \textbf{Saptarshi Ghosh\textsuperscript{1}} \\
 \textsuperscript{1} Indian Institute of Technology, Kharagpur, India \\
 \textsuperscript{2} Indian Statistical Institute, Kolkata, India \\
 \textsuperscript{3} Accenture Labs, Bangalore, India \\
}

\newcommand\blfootnote[1]{%
  \begingroup
  \renewcommand\thefootnote{}\footnote{#1}%
  \addtocounter{footnote}{-1}%
  \endgroup
}

\begin{document}
\maketitle

\blfootnote{*Equal Contribution}
\blfootnote{This work has been accepted to appear in ACL'25 findings}

\begin{abstract}

A significant portion of the energy consumed by Large Language Models (LLMs) arises from their inference processes; hence developing energy-efficient methods for inference is crucial. While several techniques exist for inference optimization, output compression remains relatively unexplored, with only a few preliminary efforts addressing this aspect. 
In this work, we first benchmark 12 decoder-only LLMs across 5 datasets, revealing that these models often produce responses that are substantially longer than necessary. 
We then conduct a comprehensive quality assessment of LLM responses, formally defining {\em six} information categories present in LLM responses. We show that LLMs often tend to include redundant or additional information besides the minimal answer. 
To address this issue of long responses by LLMs, we explore several simple and intuitive prompt-engineering strategies.
Empirical evaluation shows that appropriate prompts targeting length reduction and controlling information content can achieve significant energy optimization between 25-60\% by reducing the response length while preserving the quality of LLM responses. 
\end{abstract}

\section{Introduction}

Large Language Models~(LLMs) have seen rapid adoption in different applications, including virtual assistants for daily life, and in scientific research. However, the deployment of these models require massive computational resources, leading to significant energy consumption and carbon emissions.  
%
%
%
%
Although a single inference operation may not consume much  energy, such operations are performed enormously many times under deployment. 
For example, OpenAI's ChatGPT processes more than 1B queries per day, which consumes an estimated 0.5GWh of energy. In recent times, ninety percent of the total cloud computing demand for Amazon Web Services was for model inference purposes~\cite{barr2019amazon}.
Hence, LLM inferences contribute to a significant energy impact~\cite{wu2022sustainable,patterson2022carbon} and merit further investigation.

Our prior work established that the energy consumption during inference is closely correlated with the length of the response generated by LLMs~\cite{poddar2025benchmarking}. Notably, output length has a greater impact on inference energy usage than input length because inputs may benefit from parallel processing and caching on GPUs, whereas output generation is predominantly sequential. Additionally, we demonstrated in~\cite{poddar2025benchmarking} that the energy expended during inference is largely independent of task complexity or type, depending largely on the output size. Together, these findings highlight the potential for optimizing inference energy consumption by limiting the output length.

A key consideration in this context is how popular LLMs behave  - specifically, \textit{whether LLMs tend to generate more text than necessary for certain tasks}. 
To investigate this, in this work, we benchmark 12 LLMs across 5 tasks/datasets spanning different use cases.
We focus on factual question-answering datasets primarily consisting of factual and procedural questions, deliberately excluding opinion-based questions, as determining their minimal response length can be challenging. Our primary finding is that, \textit{for factual questions, LLM-generated responses are significantly longer than the target responses}, regardless of whether the targets are human-written or machine-generated.

It can be argued that longer answers are more useful to the users, since the additional information beyond the core information need -- such as explanations, summaries, or conversational enhancements -- can enhance user experience. 
But it is important to recognize that such enhancements come at the cost of increased energy consumption and environmental costs. A useful analogy is air travel via economy vs. business class, where the core need (timely travel) is distinct from optional enhancements that improve the experience but come at an additional cost.
This is the first work to bring forward the trade-off between energy efficiency and user satisfaction, a pivotal contribution that unveils an entirely new perspective on LLM responses. 


Next, we  categorize the LLM-generated responses by identifying six distinct information categories through a comprehensive manual survey. This categorization reveals that LLM responses often include, along with the minimal answers, explanations, examples, and reasoning information, as well as redundant or irrelevant text. 
In fact, our study reveals that, on average, the core answer constitutes only about 42\% of the total responses.
While the extra elements can enhance readability and potentially increase user satisfaction (for instance, users might appreciate supplementary context or a polite add-on), they also incur higher inference energy costs. Consequently, we argue that LLMs should be aligned to generate minimal answers by default, incorporating only a limited amount of additional content and excluding extraneous or conversational tokens.


Finally, we explore several prompt-based strategies to generate more concise responses from LLMs. These include simple directives (such as ``Answer briefly''), providing in-context examples to guide the LLM toward optimal response length, specifying a desired output length, and reducing non-essential content such as conversational and irrelevant information. 
We evaluate these prompt engineering strategies in terms of how they affect the length, energy, and information content of the LLM responses. We also check how they affect the quality of LLM responses in terms of ROUGE-L F1 scores with the target (gold standard) answers.
Empirical evaluation shows that some of our prompts achieve significant energy optimization by reducing the response lengths, while preserving (or sometimes even improving) the quality of LLM responses.
For instance, one of our best strategies is simply where we direct the LLM to provide only the minimal answer; this strategy achieves up to 88\% reduction in LLM responses, which leads up to 58\% reduction in energy consumption, while actually improving the response quality. 
\section{Related Works}
\label{sec:litsurvey}
%
The optimization of energy consumption in large language models (LLMs) has been extensively studied in the literature. Notable {white-box approaches} include pruning~\cite{kurtic2024ziplm}, quantization~\cite{xiao2023smoothquant,frantar2022gptq}, early exit strategies~\cite{schuster2022confident}, and embedding recycling~\cite{saad2022embedding}. Hardware or OS-level optimizations focus on optimal data transfer and device utilization,
for example, caching ~\cite{ramirez2023cache}, scheduling~\cite{yu2022orca}, etc. {Black-box approaches} include cascading~\cite{chen2023frugalgpt}, speculative decoding~\cite{leviathan2023fast}, input compression~\cite{jiang2023llmlingua,jiang2023longllmlingua}, among others. A detailed literature survey has been given in Appendix~\ref{app:litsurvey}.

However, the area of output compression remains relatively underexplored, with only a few preliminary efforts in the literature, namely prompt augmentation with generation directives~\cite{li2024toward}. In this work, our primary focus is to comprehensively explore the underexplored avenue of output compression for energy-efficient inference.

\section{Do LLMs generate too much text?}
\label{sec:outputlen}

In this section, we investigate the question -- \textit{Are LLMs generating too much text, more than necessary for certain types of tasks?}
We primarily focus on objective factual queries where concise and precise responses are often sufficient and compressing the response can improve energy efficiency trading-off with user satisfaction.
Note that output compression may not be suitable for long-form tasks. 


\begin{table}[!t]
    \centering
    \footnotesize
    \begin{tabular}{p{19mm}|p{48mm}}
    \toprule
    \textbf{Dataset} &
    \textbf{Description} \\
    \midrule
    \dolly\cite{DatabricksBlog2023DollyV2} & instruction following records with human-curated answers in several QA and classificaion categories\\ 
    %
    \hline
    \gooaq \cite{gooaq2021} & open-domain dataset with factual questions from Google searches and answers mined from Google's responses. \\

    
    \hline
    \msmarco \cite{DBLP:journals/corr/NguyenRSGTMD16} & Open-domain QA dataset, with questions from Bing queries and crowdsourced answers.\\ 

    

     \hline
    \narrqa \cite{kocisky-etal-2018-narrativeqa} & QA dataset where Wikipedia summaries of novels were given to humans to frame question-answer pairs. \\

    \hline
    \tweetqa \cite{xiong-etal-2019-tweetqa} & factual QA dataset, curated by humans based on context from tweets used by jounalists.\\
    
    
    \bottomrule
    \end{tabular}
    \caption{List of datasets used in our experiments.}
    \label{tab:data}
\end{table}

\begin{table*}[!ht]
    \centering
    \scriptsize
    \begin{tabular}{|p{15mm}|p{3mm}||c|c|c|c|c|p{6.5mm}|c|c|c|c|p{5mm}|p{5.5mm}|}
    \hline
        \textbf{Dataset} & \textbf{tgt} & \textbf{GPT} & \textbf{GPT} & \textbf{gemma} & \textbf{gemma}  & \textbf{Llama} & \textbf{Mistral} & \textbf{Llama} & \textbf{Llama} & \textbf{Phi-3} & \textbf{Phi-3} & \textbf{falcon} & \textbf{vicuna}\\ 
        
         & \textbf{len} & \textbf{3.5t} & \textbf{4om} & \textbf{-2-2b} & \textbf{-2-9b} & \textbf{-2-13b} & \textbf{-7B} & \textbf{-3.2-3B} & \textbf{-3.1-8B} & \textbf{-mini} & \textbf{-small} & \textbf{-11B} & \textbf{-13b} \\ \hline

        \dolly   & 68.3 & 78.2 & 123.3 & 191.1 & 152.3 & 176.3 & 150.4 & 229.0 & 539.8 & 1445.1 &  477.5 & 168.3 & 84.2 \\
        \gooaq   & 27.4 &        63.1 & 142.3 & 212.5 & 162.4 & 219.3 & 161.8 & 311.9 & 535.2 & 1333.9 &  418.5 & 177.5 & 89.0 \\
        \msmarco & 22.8 &        40.3 & 127.8 & 124.4 & 138.8 & 169.3 & 139.4 & 280.6 & 683.3 & 1620.0 & 1333.2 &  63.9 & 94.8 \\
        \narrqa  &  7.1 &        23.4 &  44.7 &  35.0 &  25.1 &  36.9 &  37.1 & 246.4 & 671.8 & 1215.6 & 1462.6 &  30.9 & 38.0 \\
        \tweetqa &  4.0 &        12.7 &  22.8 &  89.6 &  44.8 &  71.9 & 111.6 & 182.6 & 395.4 & 1671.7 & 1108.2 &  80.8 & 49.1 \\
        \hline
        
    \end{tabular}
    \caption{Average target response length (second column) vs response lengths for different LLMs on the 5 datasets. In almost all cases, LLM responses are substantially longer than the target response lengths, across all datasets.     
    }
    \label{tab:tgt_vs_gen}
\end{table*}

\subsection{Experimental Setup}
\label{sub:expsetup}



%

\noindent \textbf{Datasets:}
We choose $5$ datasets, namely \dolly, \gooaq, \msmarco, \narrqa, and \tweetqa for LLM evaluations, for reasons described below. The datasets are described in Table~\ref{tab:data} (refer Appendix~\ref{app:data_examples} for examples).

We selected these 5 datasets for the following reasons:
\textbf{(1)~Diverse tasks:}
The datasets cover a diverse range of tasks, including logical reasoning (\dolly, \gooaq), instruction following (\dolly), knowledge-dependent closed-QA~(\dolly, \gooaq), knowledge-dependent short contextual QA (\tweetqa) and long contextual open-QA (\narrqa, \msmarco).
\textbf{(2)~Diverse domains:}
These datasets also cover diverse domains, ranging from 
tweet-based queries in \tweetqa, regular factual and procedural queries from different categories in \dolly and \gooaq, and curated queries from fiction in \narrqa. \gooaq and \msmarco also present several factoid-based questions
of practical interest, sampled from google searches and bing searches. 
\textbf{(3)~Diverse answer types:}
Target answers in our selected datasets include human-generated, human-annotated, and machine-generated responses. 
To ascertain whether LLMs are generating more text than is required,  we need to compare LLM responses with minimal and sufficient answer lengths. 
Human-written answers serve as the proxy of answers of reasonable lengths in \dolly, \msmarco, \narrqa, and \tweetqa, while \gooaq contains answers to short fact-based and longer procedural questions automatically scraped from Google search answers.
This results in diverse answer types, and a rich space of answers. 




From each of the datasets, we select 1,024 samples from the validation/test splits for evaluating the LLMs. 
$5K$ samples are also extracted from the train splits to be used in Section~\ref{sec:promptengg}. To test LLMs generalizability, we use this pool of $25K$ training samples for different strategies, rather than using individual datasets for training and testing the models separately. 




\vspace{2mm}
\noindent \textbf{Large Language Models:}
In this work, we study the generation of $10$ open-source LLMs from different families --  Meta Llama-2/3, Microsoft Phi-3, Google Gemma-2, Mistral, Falcon, Vicuna. 
We also test the closed-source OpenAI GPT-3.5-turbo and GPT-4o-mini LLMs.
These models are listed in Table~\ref{tab:tgt_vs_gen} and detailed in App:\ref{app:models}. 

For reproducibility, models were loaded in standard half-precision and made to generate greedily up to a hard limit of $2K$ tokens. 
Note that this limit does \textit{not} affect the response generated, and is rather just a cut-off if the model keeps on generating tokens. 
The repetition penalty parameter was set to $1.2$~\cite{keskar2019ctrl} to restrict models from generating repetitive text.
Details of the hardware/software setup are reported in App:~\ref{app:hardware}.

\vspace{2mm}
\noindent \textbf{Measuring the length of LLM responses:}
We measure the length of the response by counting the number of tokens present in the text, where the tokenizer of the corresponding model has been used to tokenize the model output as well as the gold answer.
For the closed-source GPT-3.5/4 models from OpenAI, we use the SpaCy library to tokenize the responses and gold answers.
We refer to the length of gold-standard answers from the datasets (mostly human-written responses) as \textbf{``target length''} and the length of LLM-generated response as \textbf{``generated length''}.


\subsection{Quantitative Analysis}

\begin{figure}[!t]
    \centering	
    {\includegraphics[width=0.85\linewidth]{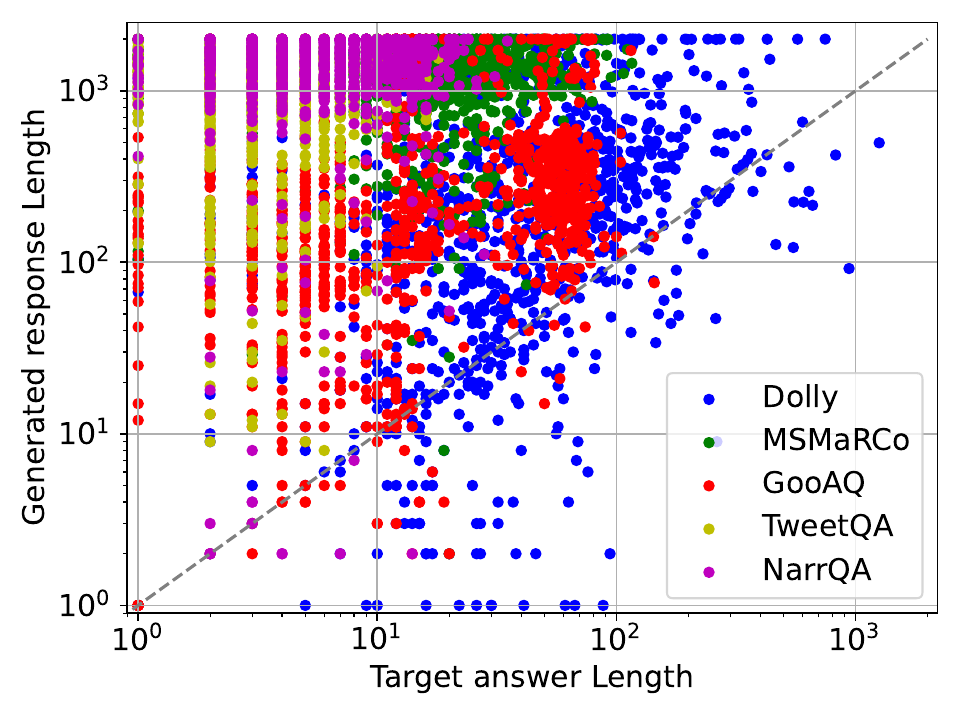}}
    \caption{Target answer length vs LLM generated response length by Phi-3-small on different datasets}
    \label{fig:tgt_vs_gen}
\end{figure}

\begin{table}[!t]
    \centering
    \scriptsize
    \begin{tabular}{|l|c|c|c|c|}
    \hline
          & \textbf{DS-R1-} & \textbf{DS-R1-} & \textbf{Gemma-2} & \textbf{Target} \\
          & \textbf{Llama-8B} & \textbf{Qwen-14B} &  \textbf{-9B} & \textbf{Answer} \\
      \hline
    \dolly & 586.2 (64.8\%) & 669.1 (73.8\%) & 152.2 & 68.3\\
    \gooaq & 552.6 (52.5\%) & 759.5 (73.1\%) & 162.4 & 27.4\\ 
    \hline
    \end{tabular}
    \caption{Average response lengths of Deepseek reasoning models on two datasets (\% of thinking/reasoning tokens inside parenthesis).}
    \label{tab:rsn_models}
\end{table}

Table~\ref{tab:tgt_vs_gen} presents the average target response length and generated response length in terms of tokens for all dataset-LLM pairs. In almost all cases, LLMs generate responses substantially longer than target responses. 
Even for extremely brief target responses (avg $<10$ tokens) in \narrqa and \tweetqa, LLM responses are relatively long (20--1.5K tokens). 

We can classify the models based on the ratio of their average response length to the average target answer length into three categories -- 
(1)~\textbf{Moderate (mostly within 1-3 times):}  models from GPT-3 family generate responses within 1-3 times of the target answer length, 
(2)~\textbf{Long (mostly between 3-10 times):} models from GPT-4, vicuna, gemma-2, llama-2, and mistral generate moderately long responses, and (3)~\textbf{Very Long (often more than 10 times):} responses of llama-3.1, and phi-3 family are generally very long. 


Figure~\ref{fig:tgt_vs_gen} plots the distribution of generated response length with target answer length across all datasets for phi-3-small; the presence of datapoints mostly above the diagonal signifies that LLMs generate overly long responses.   

Reasoning LLMs are gaining popularity as they perform well in complex scientific tasks. They do this by generating long reasoning traces for optimal final answers, requiring higher time and energy for inference. 
To further analyze the behavior of reasoning LLMs, Table~\ref{tab:rsn_models} reports average response length for DeepSeek-R1-Distill-Llama-8B and DeepSeek-R1-Distill-Qwen-14B on the Dolly and GooAQ datasets; we observe that these models generate very long outputs on average, with reasoning tokens constituting the majority of them.

Note that our selected datasets include examples with both human-written responses and standard objective responses to serve as a proxy for minimal/sufficient response length. 
Thus, we can conclude that \textit{most LLMs usually generate responses substantially longer than desired for the use cases under consideration in this study}. 


\begin{figure*}[!t]
    \centering	
    \includegraphics[width=0.8\linewidth,trim={15mm 115mm 15mm 0mm},clip]{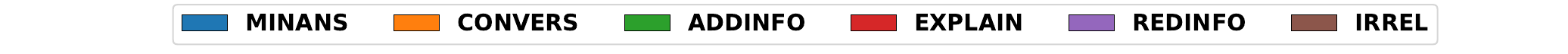}
    \vspace*{-3mm}
    
    \subfloat[Overall]{\includegraphics[width=0.165\linewidth,trim={8mm 0mm 5mm 8mm},clip]{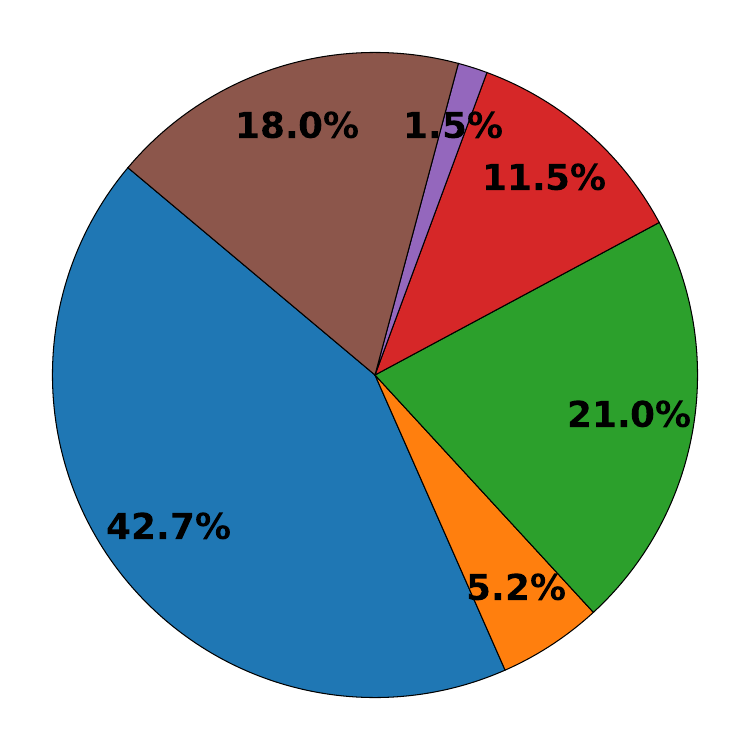}\label{fig:cat-dist-aggr}}
    \subfloat[Model-specific distribution]{\includegraphics[width=0.585\linewidth,trim={0mm 5mm 31mm 10mm},clip]{Plots_main/category/model_wise_dist.pdf}\label{fig:cat-dist-model}}
    \subfloat[Dataset-specific distribution]{\includegraphics[width=0.25\linewidth,trim={4mm 1.5mm 3mm 3mm},clip]{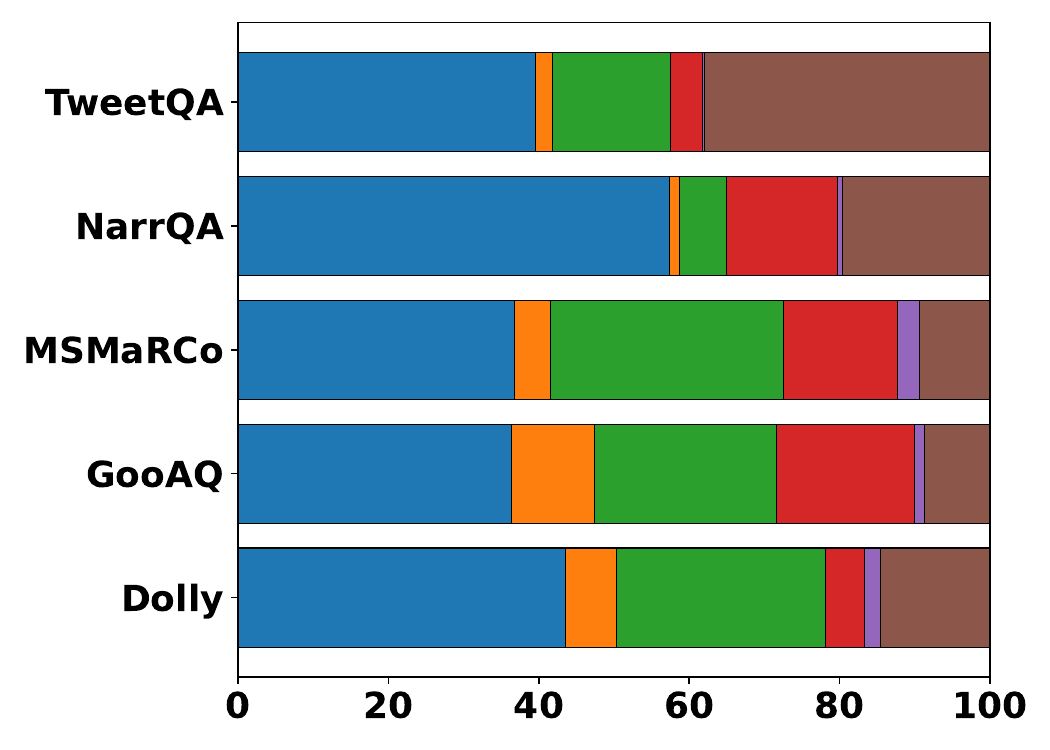}\label{fig:cat-dist-data}}
    \caption{Distribution of information categories in LLM responses.}
    \label{fig:cat-dist}
\end{figure*}

\subsection{Categorizing information in responses}
\label{sec:cat-response}

To gain further insight on overly long responses of LLMs, we categorize the information present in the responses. 
\citet{low1973role} suggests a taxonomy of irrelevant, relevant, and redundant information for general concept transfer. 
Following this line of work and our inspection of 300 responses by different LLMs, we develop the following classification scheme for the tokens contained in LLM responses. 

\vspace{1mm}
 
\noindent \textbf{1. Minimal answer~(\minans):} These tokens provide a concise and to-the-point answer to the query/prompt. Note that the minimal answer can be correct or wrong; we are \textit{not} differentiating between correct and wrong answers in this analysis.

\noindent \textbf{2. Additional helpful information~(\addinfo):} 
These include information that adds valuable context and aims to expand the knowledge of the user. Example: User: ``What is capital of France?'' LLM: ``Paris. \textit{It is a center of art and culture}''.

\noindent \textbf{3. Reasoning/Explanations~(\explain):} 
Here the LLM states its thought process, logical steps that it followed to answer the query. 
Example: User: ``What is 2 + 2?'' LLM: ``\textit{First, you take 2 and add another 2, which equals 4.} Final answer: 4''.

\noindent \textbf{4. Conversational Enhancements~(\converse):} 
Tokens that are included to maintain politeness, enhance the conversational tone, or ask users if they require further information on some aspects.
Examples include ``\textit{Let me know if you need anything else}'', ``\textit{I hope that helps!}''. 

\noindent \textbf{5. Redundant Information~(\redinfo):} In some longer responses, LLMs provide a summary of the previous content at the end. In some other cases, the responses contain rephrased versions of information already given (in the same response).
Example: ``Netherlands is the official name of the country as a whole, including Holland and other regions like Friesland, Gelderland. \textit{So, while Holland is part of the Netherlands, Netherlands is the entire country.}''



\noindent \textbf{6. Irrelevant Information~(\irrel):} At times, LLMs generate tokens unrelated to the user's query, e.g., when the model hallucinates, or produces repeating patterns of tokens. Example: User: "Who discovered America?" LLM: "Columbus! \textit{By the way, I recently read about American recipes.}"

\subsection{Analysis of LLM responses} 
\label{sub:analysis-llm-responses}


Two of the authors annotated an additional $500$ LLM responses, sampling uniformly from the $5$ datasets and $10$ LLMs, marking spans from the responses into the categories defined above. 
The first 50 were marked by both authors and achieved pairwise F-measure~\cite{deleger2012building} of 0.764, macro-averaged across different categories, which shows reasonable agreement. The major disagreements were in the REDINFO class, particularly in cases where parts of minimal answer, additional and redundant information were intertwined. Some example annotations are shown in Appendix~\ref{app:annotation}.
We are releasing this annotated dataset to promote further research on LLM responses.\footnote{\url{https://github.com/sohampoddar26/LLM-brevity}}

The distribution of the categories across all responses is given in Figure~\ref{fig:cat-dist-aggr}, whereas Figure~\ref{fig:cat-dist-model} and Figure~\ref{fig:cat-dist-data} present model-specific and dataset-specific category distributions. 
Figure~\ref{fig:cat-dist-aggr} shows that though the answers primarily contain minimal answer (MINANS) and additional information (ADDINFO), \textit{irrelevant information ($\approx 18$\%) and conversation enhancers ($\approx 5.2$\%) form a significant portion of the responses}. 
Reducing these can immediately lead to much lower consumption of energey. 
Further, restricting additional information ($\approx 21\%$) and explanations ($\approx 11.5\%$) can further compress the answers without degrading the quality, where concise answers are sufficient.

Figure~\ref{fig:cat-dist-model} shows the distribution of information categories in the responses by different LLMs. 
We see that, for some LLMs such as GPT-3,4, Mistral-7B, gemma-2, Llama-3.2-3B, most of the generated tokens belong to `minimal answer' with a moderate portion of additional information. 
On the other hand, llama-3 generates a lot of explanation (CoT reasoning) and irrelevant information, and phi-3 generates a high amount of irrelevant text, with minimal response being rather small parts of the generations.  
These model-specific trends in responses seem to be consistent across a model family, suggesting that the training setup of an LLM is the primary factor in determining the distribution of content categories within a response in comparison to the relative size of the model. 
Interestingly, \textit{older LLMs (Llama-2.1, GPT-3.5) give more minimal answers, whereas newer LLMs (Llama-3.1, Llama-3.2, GPT-4om), with updated pre-training data and strategies, generate more of additional information and explanations.} 

For dataset-specific trends (Figure~\ref{fig:cat-dist-data}),  queries in \narrqa and \tweetqa are objective and factual, based on a given context, resulting in responses being mostly minimal answers with low amount of additional information. 
In \msmarco, \dolly, and \gooaq, queries are open-domain questions, containing both factual and reasoning queries, resulting in decreasing fraction of minimal responses and increasing additional info and conversational enhancements. 

Based on our observations above, we now explore several strategies to reduce the LLM response lengths, while still retaining the response quality. 
%
%


\section{Some common strategies for shortening LLM responses}

We now focus on some common strategies for shortening LLM responses to achieve energy-efficient inference. We explore simple, intuitive white-box approaches, including response truncation and fine-tuning of LLMs, as well as black-box strategies based on prompt engineering. 

\textbf{Truncating responses} refers to stopping the generation early, e,g., just after the minimal answer is obtained. 
However, we notice that 
in about 19\% of the responses studied in Section~\ref{sub:analysis-llm-responses}, the minimal answer is \textit{not} at the beginning of the LLM response. 
In most of these cases, the LLMs output conversational enhancers first to set up the context or for readability, followed by the minimal answer, thus limiting the scope of strategies for truncating responses. 

\vspace{2mm}
\noindent \textbf{Fine-tuning LLMs:} 
Next we explore whether simple fine-tuning can help in controlling the response length. Toward that, we performed some basic fine-tuning experiments.  
We fine-tuned Phi-3-mini (one of the LLMs that generate very long responses, as stated earlier) with a pool of human-written question-answers from the \dolly and \gooaq datasets, using LoRA with $r=16$, for $100$ epochs. The models were trained on the target answer tokens with the hypothesis that it would enable the models to generate responses similar to humans on these particular datasets. 
We compare the average ratio of generated and target response lengths for the \default  model and the fine-tuned model. We observe that \textit{simple fine-tuning is leading to 1.24 times higher ratio for \dolly and 1.26 times higher ratio for \gooaq, in comparison with the \default model}. 
This is an unexpected result, where fine-tuning is (slightly) increasing the response length instead of shortening the length. 
This behavior could stem from extensive pretraining on large-scale data, making small-scale fine-tuning insufficient to alter its learned patterns.
Additionally, a model's intrinsic stopping mechanism may not be functioning as intended after the fine-tuning, causing it to overestimate the length for the responses. 
%

Therefore, it is prudent to explore prompt engineering strategies leveraging LLM's zero-shot capabilities, drawing on our observations in Section~\ref{sec:outputlen}. We explore this direction in the next section.

\section{Prompt engineering strategies for controlling LLM responses}
\label{sec:promptengg}


Here we investigate various prompt engineering strategies 
aimed at generating concise LLM responses, thereby reducing inference energy. 


\subsection{Prompt engineering strategies}

We first describe the different strategies we used to control response length. The exact prompts used are given in Table~\ref{tab:prompts}.

\vspace{1mm}
\noindent \textbf{\brief:} Following \cite{li2024toward}, the default query is modified by appending ``\textit{Answer briefly}'' at the end of the query. 


\vspace{1mm}
\noindent \textbf{\incontxt:} In this method, we provide a set of in-context examples to help LLM in deciding the ideal response length of the given query. Specifically, we provide a sequence of 10 sample queries and their respective gold-answers as in-context examples, followed by the given query. These samples were chosen from the $25K$ train set based on the similarity with the given query, as predicted by the BM25+ retrieval model~\cite{trotman2014improvements}. 


\vspace{1mm}
\noindent \textbf{\limit:}  
In this approach, we specify an estimated length of the response and instruct the LLM to restrict the generation length within that limit. Specifically, we add ``\textit{Answer within $\mathcal{X}$ words}'' at the end of the prompt, 
where $\mathcal{X}$ is the limit of generation that we obtain through three ways:
\textbf{(1) \similarbm :} Top-k similar samples are selected from the pool of samples using BM25+ model, and $\mathcal{X}$ is set to the median of their target lengths. 
\textbf{(2) \goldreslen :} To serve as an oracle baseline, $\mathcal{X}$ is set to the actual gold-answer length corresponding to the query.
\textbf{(3) \predreslen:} Here we employ a supervised framework for learning the ideal length of a given query. Specifically, we train an encoder-only transformer-based regression model \debert~\cite{he2021deberta} with scaler output to predict $\mathcal{X}$ of the given query. The model is trained with queries and the length of their target (gold standard) responses from the training set. 

\vspace{1mm}
\noindent \textbf{\limitcat:}  
Based on our observations regarding the distribution of various categories in LLM responses (see Figure~\ref{fig:cat-dist-aggr}), we hypothesize that responses can be made shorter and energy-efficient while preserving response quality, by reducing the presence of \converse, \redinfo, and \irrel in the responses.
Accordingly, we design {\limitcat}, a set of prompt directives to control the presence of various categories in the response. 
We tried several combinations of prompts to restrict the category distribution in the responses. Out of these we report results for the following two strategies which performed the best: 
\textbf{(1)~\mans:} Model is asked to provide the minimal answer only. 
\textbf{(2)~\mansred:} Model is asked to provide the minimal answer, along with some additional information if applicable. It is also explicitly asked \textit{not} to produce redundant text or conversational enhancements
(exact prompts in Table~\ref{tab:prompts}).

\begin{table}[tb]
    \centering
    \footnotesize
    \begin{tabular}{p{20mm}|p{45mm}}
    \hline
    query & <\textit{Context if present}> \textbf{\textbackslash n} \\ 
    &     \textbf{\#\#\# Question:} <\textit{Query text}> \textbf{\textbackslash n}  \\
    prompt & <\textit{query}> <\textit{response directive}> \\
    & \textbf{\#\#\# Response:}\\
    \hline \hline
    
    \multicolumn{2}{l}{\textbf{Response directives}} \\
    \hline
    \brief & Answer briefly. \\
    \mans & Only provide the minimal answer. \\
    \mansred & Provide the minimal answer. Provide some additional information where applicable, but do not produce redundant text or politeness and conversational enhancements.\\
    \limit & Answer within $X$ words. \\
    \hline \hline
    \incontxt & <\textit{10 examples}> <\textit{default prompt}> \\
    \hline
    
    \end{tabular}
    \caption{Prompt directives to shorten LLM responses.}
    \label{tab:prompts}
\end{table}

\subsection{Results}
We empirically validate the efficacy of the prompt engineering strategies against the default prompt, by investigating these questions: 
(1)~Can the prompt engineering strategies achieve a reduction in LLM response length? 
(2)~Does the length reduction degrade the response quality?
(3)~Are these approaches able to achieve notable reduction in terms of energy?
(4)~How do these methods influence the category distribution in the LLM responses?

\begin{figure}[!t]
    \centering	

    \includegraphics[width=\linewidth]{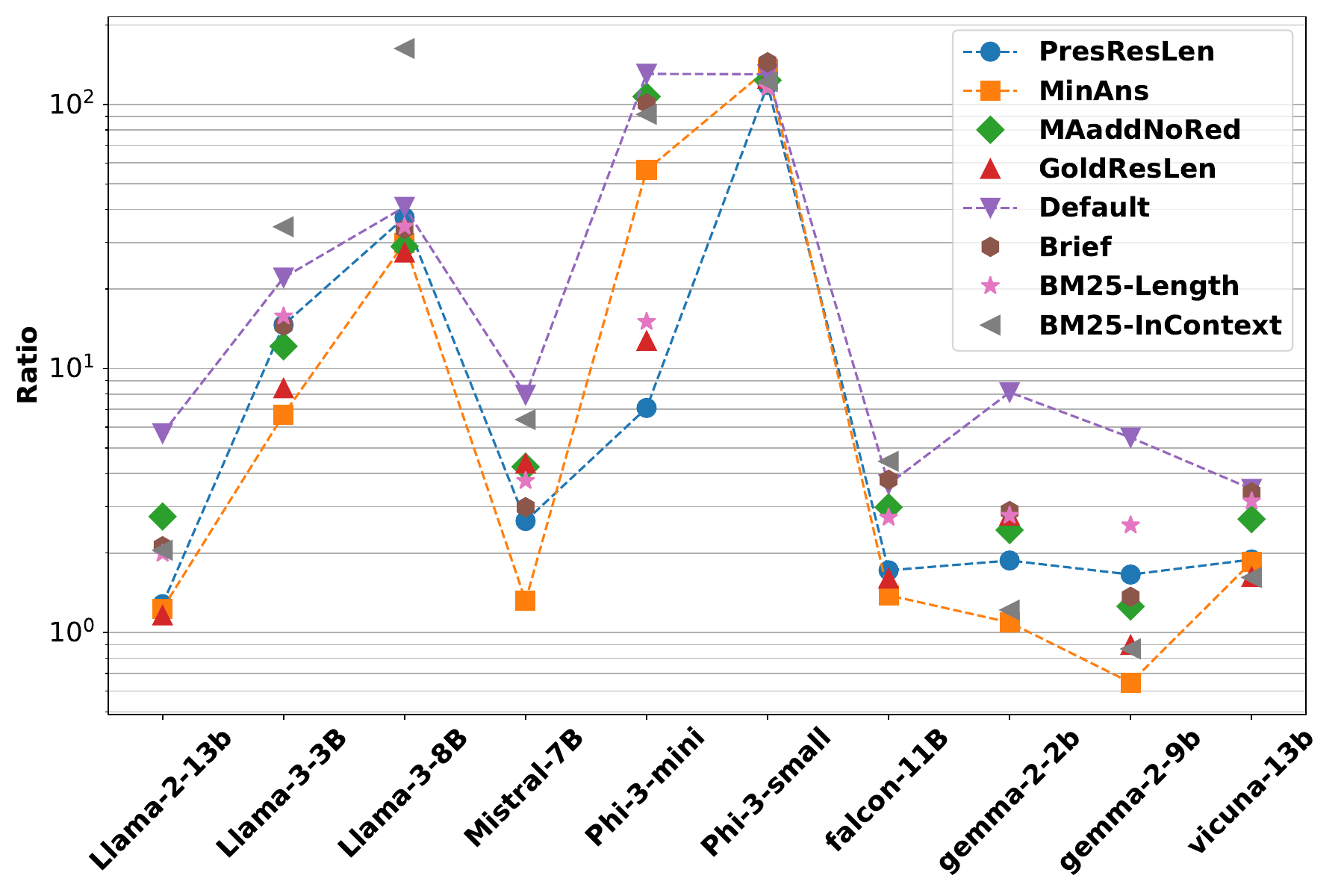}
   \caption{Ratio of generated response length to target length for different LLMs using each prompting strategy, averaged over all datasets. 
   }
    \label{fig:ratio}
\end{figure}

\vspace{2mm}
\noindent \underline{\textbf{Effects on response length:}}
Fig.~\ref{fig:ratio} reports the ratio of \textit{generated response length} to the \textit{target length} for each of the above-mentioned strategies, averaged across all datasets. 
We observe that prompt directives are generally able to reduce the response length for almost all models relative to the \default setup. 
%
%

\mans achieves the highest length reduction across all models with about 60\% decrease, followed by \predreslen at 53\%. 
The supervised length estimator \predreslen performs at par with the oracle (\goldreslen) with superior length reduction for some models, namely mistral~(66\%), llama-2~(77\%), and gemma-2-9b~(69\%), showing its efficacy in accurate estimation of ideal target length.  
\similarbm and \brief perform comparably while achieving lower compression of about 38\%. on average 
Length reduction is minimal for \incontxt, as some models fail to understand the desired output length from examples.
\mansred produces relatively less compression too as we specifically ask the models to generate additional information.
However, some LLMs, like phi-3-small or llama-3-8B fail to respond to such prompts appropriately.

\vspace{1mm}
\noindent \underline{\textbf{Effect on response quality:}}
While response quality is abstract and difficult to quantify, we wanted to check if the user's 
\textit{basic information need} is being fulfilled in the LLM responses.
To this end, we compute the Rouge-L F1 scores of the LLM responses with respect to the target answers. 
While this is no substitute for a human evaluation, this will give us some insights into the quality of responses using different strategies.

%

\begin{figure}
    \centering	
    {\includegraphics[width=\linewidth]{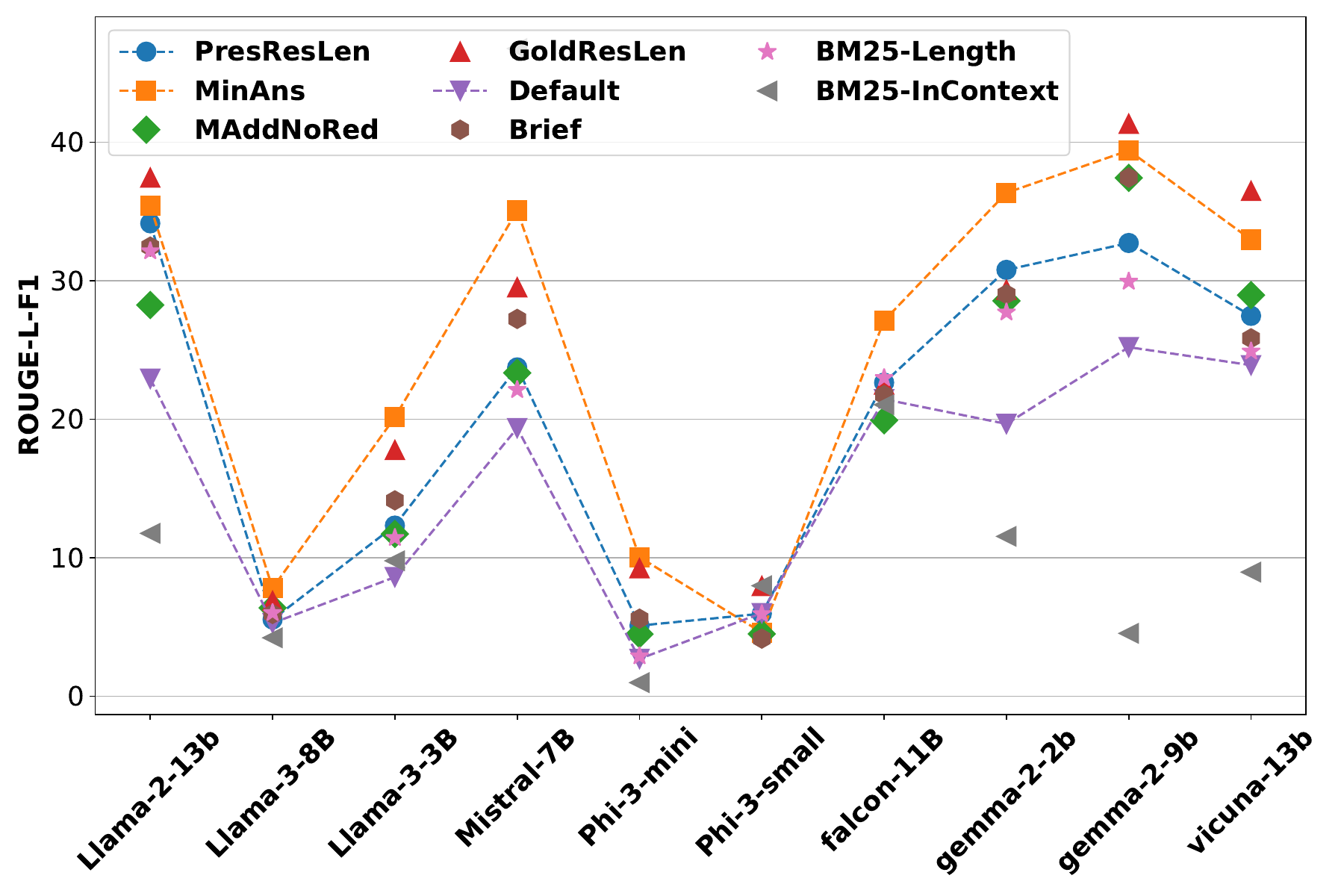}}\hspace{6mm}
    \caption{ROUGE-L-F1 scores for different models using different strategies, averaged across all datasets.}
    \label{fig:accuracy}
\end{figure}

Fig.~\ref{fig:accuracy} shows the Rouge-L-F1 score of each strategy for different models, averaged across all datasets. 
We observe that employing the above-mentioned prompting strategies generally improves the F1-scores of the responses upon the default setup in most cases, since the models are instructed to generate more information-dense answers, that better match the target answers. 
Specifically, we find that \textit{shorter responses lead to substantial improvements in ROUGE precision, with marginal drop in ROUGE recall, leading to increased F1 scores} (
refer to Appendix~\ref{app:rouge} for the precision, recall and F1-scores for all the methods).

%
\mans emerges as the most effective technique with the highest gain in response quality,  followed by \goldreslen, \predreslen and \brief.
These results demonstrate that both simple directives for shortening responses (\mans, \brief) as well as providing accurate length estimates (\predreslen) can achieve significant length reduction while preserving the response quality. 
\mansred and \similarbm shows a smaller performance gain than the default, which can be attributed to longer response with lesser precision.
\incontxt generates even larger responses leading to drop in performance. 

   



\begin{figure}
    \centering	
    \includegraphics[width=0.99\linewidth]{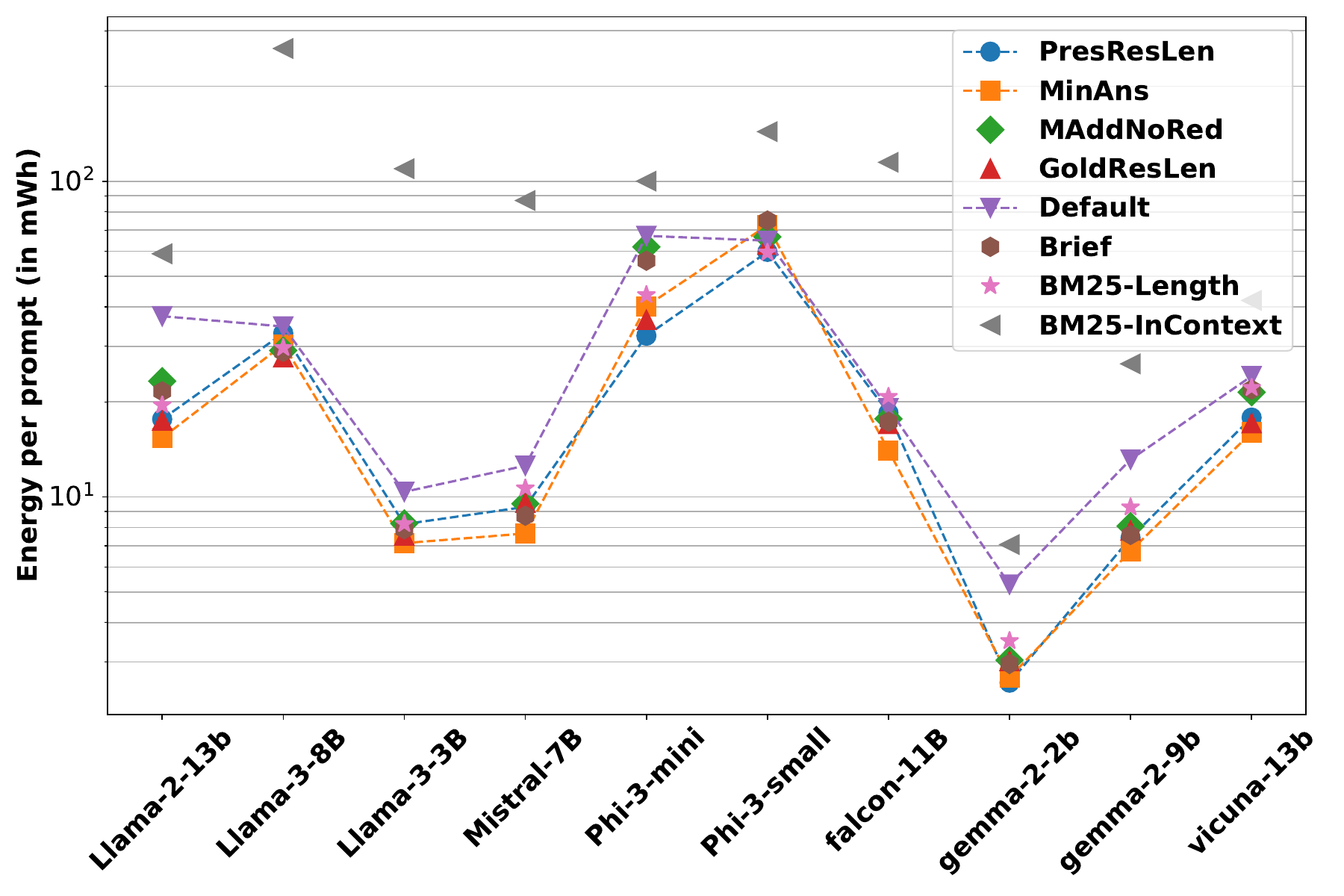}
   \caption{Inference Energy for different models using different strategies, averaged over all datasets }
    \label{fig:energy}
\end{figure}

\vspace{1mm}
\noindent \underline{\textbf{Effects on Energy Consumption:}}
To understand the effect of prompt engineering on inference energy, we report inference energy for different models for different strategies, averaged across all datasets in Figure~\ref{fig:energy}. The inference energy consumed per sample is measured in milli-watt-hour (mWh) using the popular CodeCarbon library~\cite{codecarbon} (see Appendix~\ref{app:hardware}). 

%

Most methods demonstrate improved energy efficiency compared to the default approach, except for \incontxt, which fails to reduce energy usage across all models, attributing to both significantly longer input and its ineffectiveness in achieving length reduction.
Among the evaluated techniques, \mans emerges as most energy-efficient method with 28\%, attributing to highest length reduction. 
\mans is closely followed by \predreslen (comparable with oracle \goldreslen) with about 26\% reduction, highlighting the importance of advanced energy-efficient supervised approaches for estimating optimal response lengths.
Methods like \similarbm achieves lower energy efficiency, due to its compression inefficiency and computational overhead. 

\begin{table*}[!t]
    \centering
    \scriptsize
    \begin{tabular}{|l|p{135mm}|}
    \hline
    \textbf{Strategy} & \textbf{Gemma-2-9B Response}  \\
    \hline
    \multicolumn{2}{|c|}{\bf Query: 29 cm is equal to how many m?} \\
    \hline
    \default & Here's how to convert centimeters (cm) to meters (m): **Remember the conversion:**  1 meter = 100 centimeters.
    **Divide by 100:** To find out how many meters are in 29 cm, divide 29 by 100.
    29 cm / 100 cm/meter = 0.29 meters Therefore, 29 cm is equal to 0.29 meters \\
    \hline
    \mans & 0.29 \\
    \hline
    \brief & 0.29 m\\
    \hline
    \mansred & 0.29 
    **Explanation:** There are 100 centimeters (cm) in one meter (m). To convert from centimeters to meters, divide by 100." \\
    \hline
    \multicolumn{2}{|c|}{\textbf{Query: \texttt{<INFORMATION from different sources>} \#\#\# Question: what is a necromonger in riddick?}} \\
    \hline
    \default & In the Chronicles of Riddick film franchise, Necromongers are a fanatical warrior-religious order ... \texttt{<LONG ADDITIONAL INFO>} Let me know if you have any other questions about the Chronicles of Riddick! \\ 
    \hline
    \mans & A fanatical religious warrior race seeking universal conversion or death. \\
    \hline
    \brief & A Necromonger is a member of a fanatical religious empire in the *Riddick* franchise. They believe life is an enemy of the universe and seek to conquer and convert all others through violence. \\ 
    \hline
    \mansred & A Necromonger is a member of a fanatical religious empire in the *Chronicles of Riddick* franchise.
    They believe life is an affront to the universe and seek to eradicate it through conversion or death. Their society is highly structured with strict rules and rituals.  \\
    \hline \hline
    
    \multicolumn{2}{|c|}{\bf Query: How to connect iphone to LG TV to play music?} \\
    \hline
    \default & 
    Here's how to connect your iPhone to an LG TV for screen mirroring, along with some important things to keep in mind: \texttt{<LONG ADDITIONAL INFO>} Let me know if you have any other questions! \\
    \hline
    \mans & Use AirPlay 2 on your iPhone and ensure your LG TV supports it. Then, select ""Screen Mirroring"" in Control Center and choose your LG TV from the list. \\
    \hline
    \brief & 1. **Enable Miracast on your LG TV.** (Settings > General > Screen Share) 
    2. **Turn on AirPlay Mirroring on your iPhone.** (Control Center > Screen Mirroring).
    3. **Select your LG TV from the list of available devices.**
    Let me know if you need detailed instructions! \\
    \hline
    \mansred & 1. **Enable Miracast on your LG TV.** (Settings > General > Screen Share) 
    2. **Turn on AirPlay Mirroring on your iPhone.** (Control Center).
    Your iPhone should now appear as a connection option on your LG TV. Select it to mirror your screen. \\
    \hline
    
    \end{tabular}
    \caption{Sample queries and responses of Gemma-2-9B after introducing prompt strategies.}
    \label{tab:examples}
\end{table*}

\vspace{1mm}
\noindent \underline{\textbf{Effects on information categories in responses:}} Finally, to understand how these prompts affect the information category distribution in the LLM responses, we perform the following experiment. We select a subset of the category-annotated queries for which the responses were manually annotated to identify information categories (Sec.~\ref{sub:analysis-llm-responses}). 
For each of the queries, we employ the following prompts -- \brief among the general prompts, \predreslen among the length-specific prompts, and \mans, and \mansred among the category-specific prompts. 
The responses of the two models on these four prompts are further annotated by the authors into the same information categories.

\begin{figure}[!t]
    \centering	
    \includegraphics[width=\linewidth,trim={50mm 115mm 35mm 0mm},clip]{Plots_main/category/model_wise_dist.pdf}
    \vspace*{-8mm}
    
    \subfloat{\includegraphics[width=\linewidth,trim={4mm 5mm 31mm 15mm},clip]{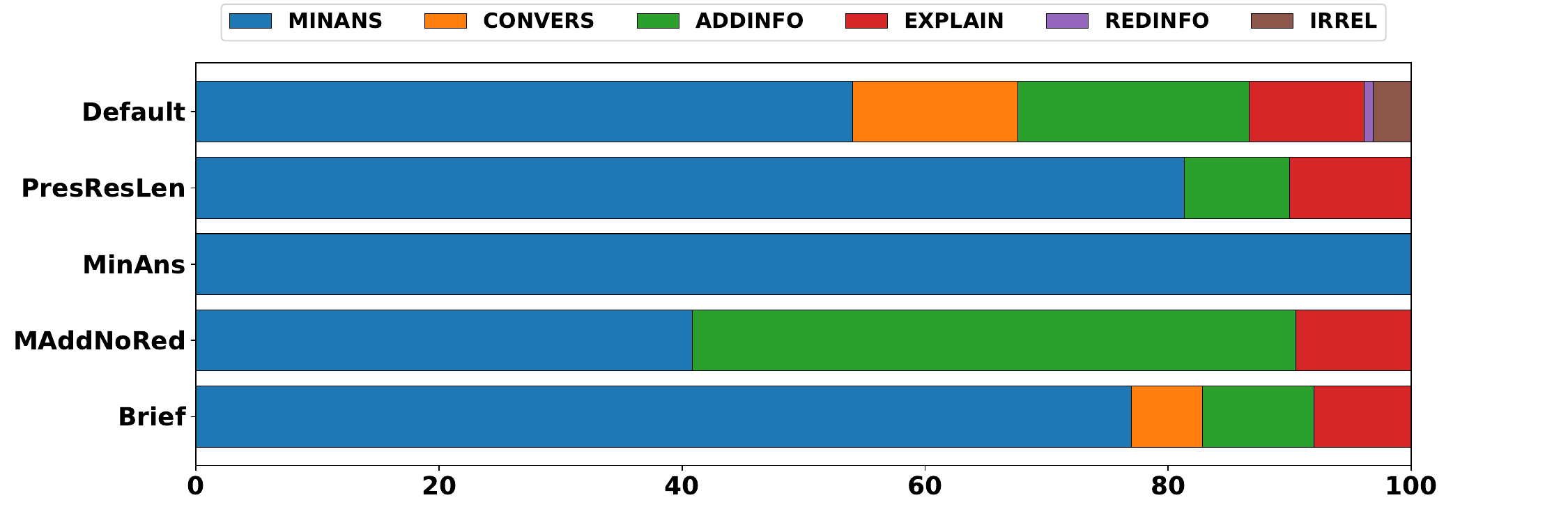}}
    
    
    
    \caption{Distribution of information categories in responses of Gemma-2-9B after incorporating 4 prompt strategies, compared to \default.}
    \label{fig:cat-dist-prompt}
\end{figure}

Figure~\ref{fig:cat-dist-prompt} presents a representative result for the gemma-2-9B LLM, showing the category distribution for the generated responses, averaged across all queries for each model.
Gemma-2 is seen to give almost entirely minimal answers for \mans, and a good balance of minimal and additional information with the \mansred strategy. \brief and \predreslen give relatively short answers containing mostly minimal answers, with \brief having some amounts of all desirable classes. 

Table~\ref{tab:examples} shows a few examples of responses by Gemma-2-9B before and after introducing the prompt strategies.
It can be seen that \minans leads the Gemma-2-9B model to generate just the minimal answer and correctly remove the explanations. 
We also see examples for \brief and \mansred which shorten the responses hugely, while still retaining some useful extra information.
However, we also observed cases where the prompts are making unnecessary cuts. 
For example, as stated earlier, we experimented with different \limitcat strategies; we observed that, if we ask the  Gemma-2-9B model to remove the conversation elements, there are cases where it also removes explanations and additional info as well, leading to a more minimal answer that may affect user satisfaction.
This suggests that further research is needed to systematically identify prompts that lead to the desired behaviour (with respect to both response length and quality) in various LLMs.

\section{Concluding Discussion}

In this work, we comprehensively explore output compression for energy-efficient inference. Our specific contributions are as follows. 
(1)~We are the first to systematically establish that large language models (LLMs) often generate very long responses which lead to high energy consumptions; We also categorize LLMs based on their generation lengths. 
(2)~We present the first comprehensive information categorization of the LLM responses, identifying $6$ information classes; 
(3)~Based on our findings, we develop various length-based and category-based prompts and evaluate them extensively in terms of how they affect the length, energy and quality of LLM responses. We observe that appropriate prompts can achieve significant energy reduction by reducing response length, alongside improving response quality. 
Overall, we think the present work takes the first important step toward exploring the utility of output compression in reducing the energy consumption of LLMs.

Our work opens up several future directions for further work (more details in Appendix~\ref{app:futurework}). 
\noindent \textbf{(1)~What constitutes an ideal response length} under different scenarios/applications should be studied more, and various factors influencing that decision should be identified. 
Measuring user satisfaction would likely require different approachs, including human surveys and automated scoring metrics and need a study on its own. 
\textbf{(2)~User personalisation:}
Another possible direction is to consider user-preferences while optimizing response length, since different users can prefer responses of different lengths for the same question/task. Such personalization is out of scope for the current work. 
However, it presents a promising line of future research. 
\textbf{(3)~Automatic information categorization:} We attempted to automate the categorization of LLM responses. To this end, we employed OpenAI GPT-4o with a detailed prompt describing all the information categories. We achieved an average F1 score of 0.391 over all tokens. Hence, efficient methods for distinguishing categories in LLM responses  should be investigated. 
\textbf{(4)~Pre-training:} Model families seem to generate responses with similar category distributions, indicating the pre-training of the LLMs should be revisited to have better control over what information categories should be included in the response. 
\textbf{(5)~Model alignment:} 
Efforts should be made to align LLMs towards generating responses of desirable lengths. 
%
\textbf{(6)~Efficient decoding algorithms} may be employed to control category information within the generated responses in the real-time, to achieve a better compromise between readability and energy-optimality. 

\section*{Acknowledgments}

The authors thank the anonymous reviewers whose suggestions helped to improve the work.
The research is partially supported by a research grant from Accenture Corporation.
Paramita Koley is supported by SERB NPDF Fellowship.

\section*{Limitations}
While this study offers a comprehensive analysis and valuable insights, some limitations should be noted:  
(1)~Limited applicability to long-answer tasks: The potential for energy optimization through output compression is limited in tasks requiring detailed and lengthy responses, such as creative writing or critical decision-making tasks with explanations. Our approaches are primarily effective for objective, factual queries where concise and brief responses are both sufficient and desirable.  
(2)~Subjectivity of Ideal Response Length: The notion of an ideal response length is subjective and varies significantly among individuals. As a result, a one-size-fits-all approach may not effectively meet the diverse needs of users. Here we argue that the default behaviour of LLMs can be tuned so that they generate shorter responses, so that energy consumption is minimized. A user can ask for more details or longer responses if she desires. 
(3)~Use of standard datasets: Our experiments have been performed on publicly available rich data for testing LLMs. However, the LLMs we used might have been pre-trained on parts of these standard datasets, causing them to behave in a certain way. This may lead to slight variations in results for real-world use cases, but the overall findings should be generalizable enough.

\bibliography{custom}

\newpage

\noindent \textbf{\large Appendix}

\appendix

\section{Related Works}
\label{app:litsurvey}
%
The optimization of energy consumption in large language models (LLMs) has been extensively studied in the literature.

\noindent \textbf{White-box approaches:} These techniques focus on architectural modifications to LLMs and represent a well-established line of research. Notable methods include pruning~\cite{kurtic2024ziplm}, quantization~\cite{xiao2023smoothquant,frantar2022gptq}, early exit strategies~\cite{schuster2022confident}, and embedding recycling~\cite{saad2022embedding}, among others.  

\noindent \textbf{Hardware- or OS-level optimizations} focus on improving energy efficiency by leveraging hardware innovations or operating system enhancements. 
Yu et al.~\cite{yu2022orca} propose an optimal scheduling strategy by making it iteration-level from user-request-level. 
Flash-attention~\cite{dao2022flashattention} effectively reduces latency by optimizing the number of memory read and write operations required in calculating model states.
Ramirez et al.~\cite{ramirez2023cache} employs a student model along with the larger LLM, which gradually gains proficiency in independently handling increasing user requests, mimicking the popular caching technique in OS. 
 
\noindent \textbf{Black-box approaches} avoid altering the underlying model, unlike the white-box models. Instead, they employ multi-level frameworks that incorporate smaller LLMs for specific tasks. Chen et al.~\cite{chen2023frugalgpt} highlight the efficacy of cascading and multiplexing in reducing inference energy and cost. Leviathan et al.~\cite{leviathan2023fast} introduce speculative decoding that accelerates the decoding by running Larger LLMs on the output of approximately smaller models. 

\noindent \textbf{Input compression} techniques apply for large-input tasks like legal question answering, multi-document summarization, multi-document retrieval, etc, where the input is excessively high. Jiang et al.~\cite{jiang2023llmlingua} propose LLMLingua, a coarse-to-fine prompt compression method that involves a budget controller to maintain semantic integrity under high compression ratios. LongLLMLingua~\cite{,jiang2023longllmlingua} further refine the method, resulting in achieving higher compression. 


As already shown in literature~\cite{luccioni2024power,poddar2025benchmarking}, inference energy is also closely tied to output length, presenting a scope for reducing energy consumption by \textbf{Output compression}. However, this area remains relatively underexplored, with only a few preliminary efforts in the literature. For instance, prompt augmentation with generation directives has been investigated as an initial step~\cite{li2024toward}.  In this work, we aim to comprehensively explore the underexplored avenue of output compression, advancing the state of the art in energy-efficient inference.  






\section{Details of LLMs used}
\label{app:models}

Table~\ref{tab:models-app} states the links to the specific LLM variants that we used in our experiments.


\section{Details of Hardware/Software setup}
\label{app:hardware}

We perform our experiments on a single NVIDIA A6000 GPU with 48GB VRAM hosted in a local server with Intel Xeon Silver 4210R processor and 128GB RAM, running Ubuntu 20.04-LTS. We use Pytorch version 2.3 (with CUDA 12.1). We employ vLLM~\cite{kwon2023efficient} for all our experiments, which is a fast and easy-to-use library for LLM inference and serving.

To measure energy consumption of LLMs in offline mode, we used the CodeCarbon~\cite{codecarbon} library with default settings. The library samples every 15 seconds the GPU power usage from the \textit{nvidia-smi} app, CPU-power from Intel RAPL files and an approximate RAM power. These power values are integrated over time to get the total energy usage.
Note that CodeCarbon is a popular and reliable tool for measuring energy consumption~\cite{jay2023experimental,bouza2023estimate} and has been used in prior works to measure the energy consumption of LLMs~\cite{poddar2025benchmarking}.

\begin{table*}[!ht]
    \centering
    \footnotesize
    \begin{tabular}{l|p{110mm}}
    \toprule
    Model & Model description link\\
    \midrule
        \textbf{GPT-3.5-turbo} \& \textbf{GPT-4o-mini} & https://platform.openai.com/ \\
        \hline
        \textbf{Llama-2-13B}~(13B params) & https://huggingface.co/meta-llama/Llama-2-13b-chat-hf \\ 
        \textbf{Llama-3-3B}~(3B params) & https://huggingface.co/meta-llama/Llama-3.2-3B-Instruct \\
        \textbf{Llama-3-8B}~(8B params) & 
        https://huggingface.co/meta-llama/Llama-3.1-8B-Instruct \\
        \hline
        \textbf{Gemma-2-2B}~(2.6B params) &  https://huggingface.co/google/gemma-2-2b-it \\
        \textbf{Gemma-2-9B}~(9B params) &  https://huggingface.co/google/gemma-2-9b-it \\
        \hline
        \textbf{Phi-3-mini}~(4B params) & https://huggingface.co/microsoft/Phi-3-mini-4k-instruct \\
        \textbf{Phi-3-small}~(8B params) & https://huggingface.co/microsoft/Phi-3-small-8k-instruct \\
        \hline
        \textbf{Mistral-7B}~(7B params) & https://huggingface.co/mistralai/Mistral-7B-Instruct-v0.3 \\
        \textbf{Falcon-11B}~(11B params) & https://huggingface.co/tiiuae/falcon-11B \\
        \textbf{Vicuna-13B}~(13B params) & https://huggingface.co/lmsys/vicuna-13b-v1.5 \\
    \bottomrule
    \end{tabular}
    \caption{Links to specific models versions we used in our experiments}
    \label{tab:models-app}
\end{table*}

\begin{figure*}[!ht]
    \centering	
    \subfloat[Llama-3.1-8B]{\includegraphics[width=0.48\linewidth]{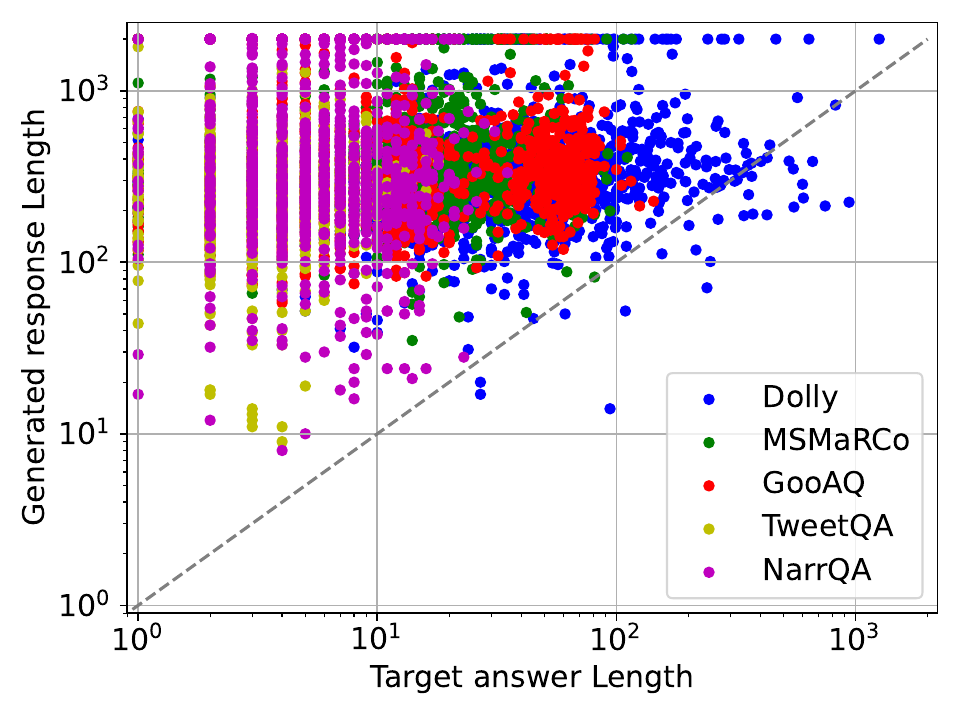}}
    \subfloat[Gemma-2-9B]{\includegraphics[width=0.48\linewidth]{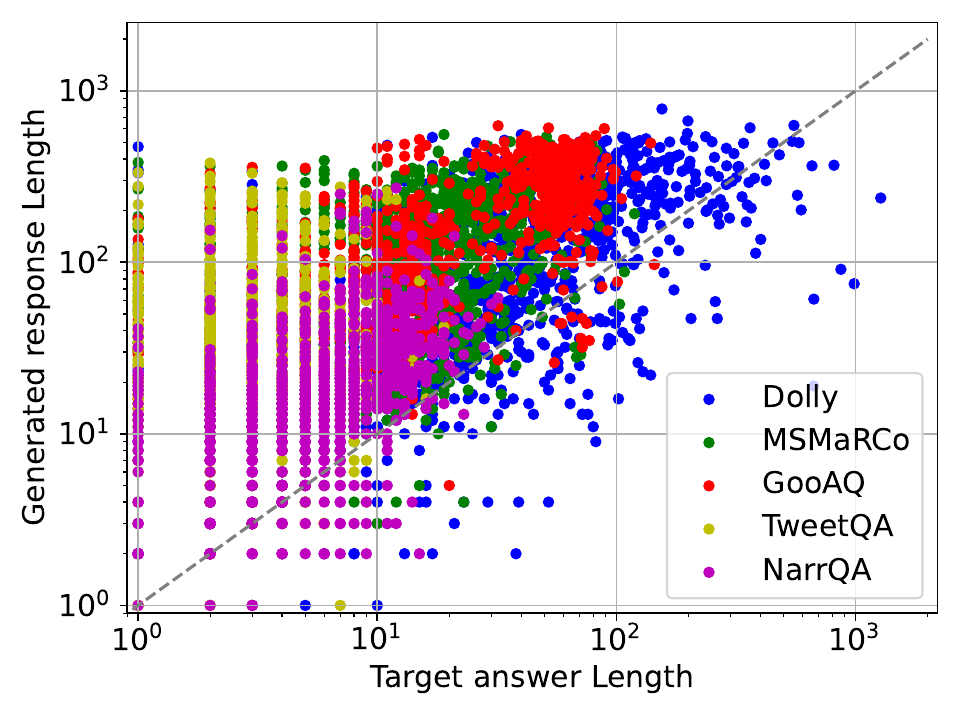}}
    \caption{Target answer length vs LLM generated response length on different datasets}
    \label{figapp:tgt_vs_gen}
\end{figure*}


\section{More Length ratio plots}
\label{app:ratio}

Figure~\ref{figapp:tgt_vs_gen} shows scatter plots comparing the target answer lengths and the generated response lengths for some more LLMs. In all cases, we see that LLMs generate substantially longer responses then the target answer lengths, in most cases. 


\section{Annotation Examples}
\label{app:annotation}

\begin{figure*}[!ht]
    \centering	
    {\includegraphics[width=0.9\linewidth]{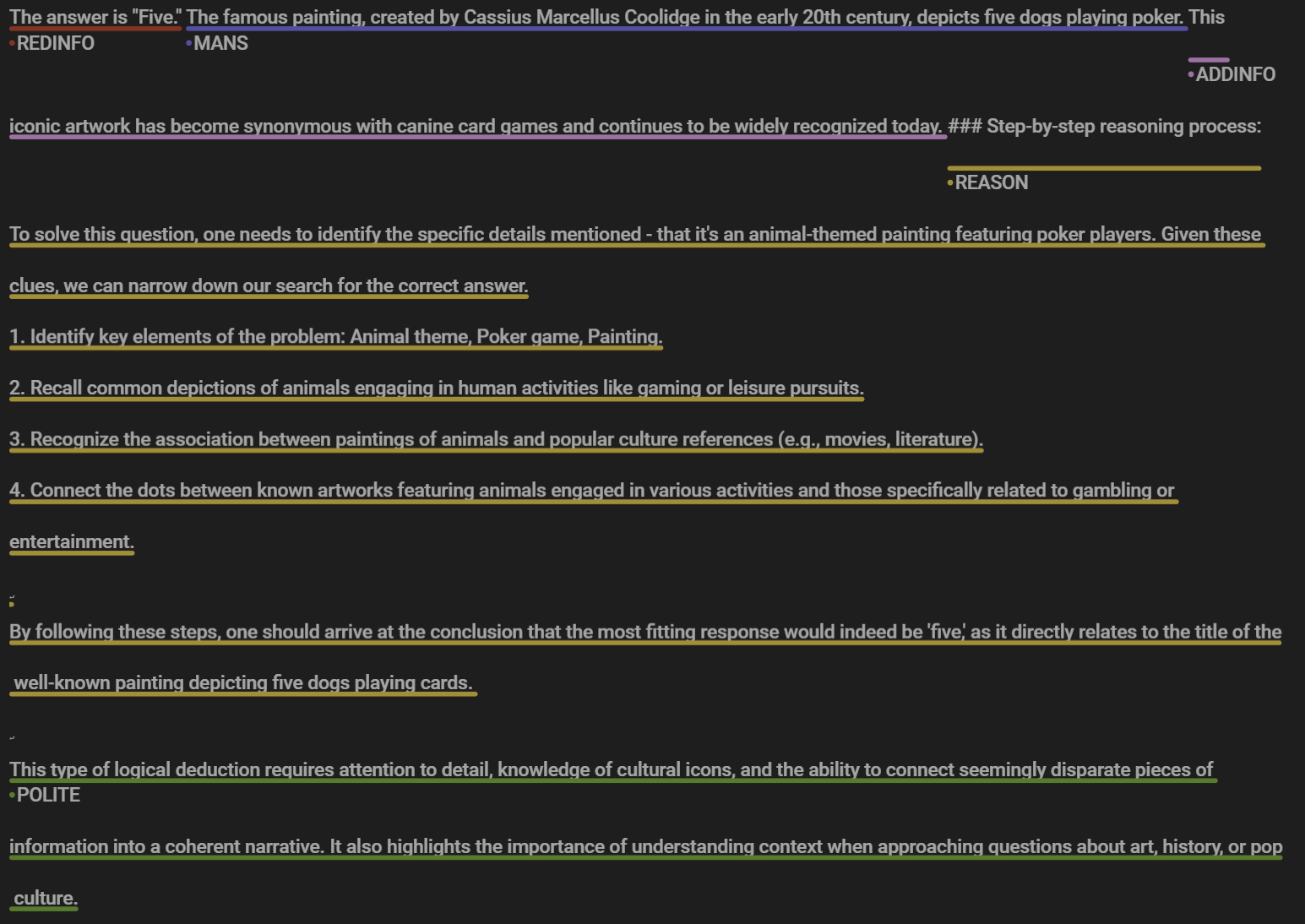}}

    \vspace{5mm}
    
    {\includegraphics[width=0.9\linewidth]{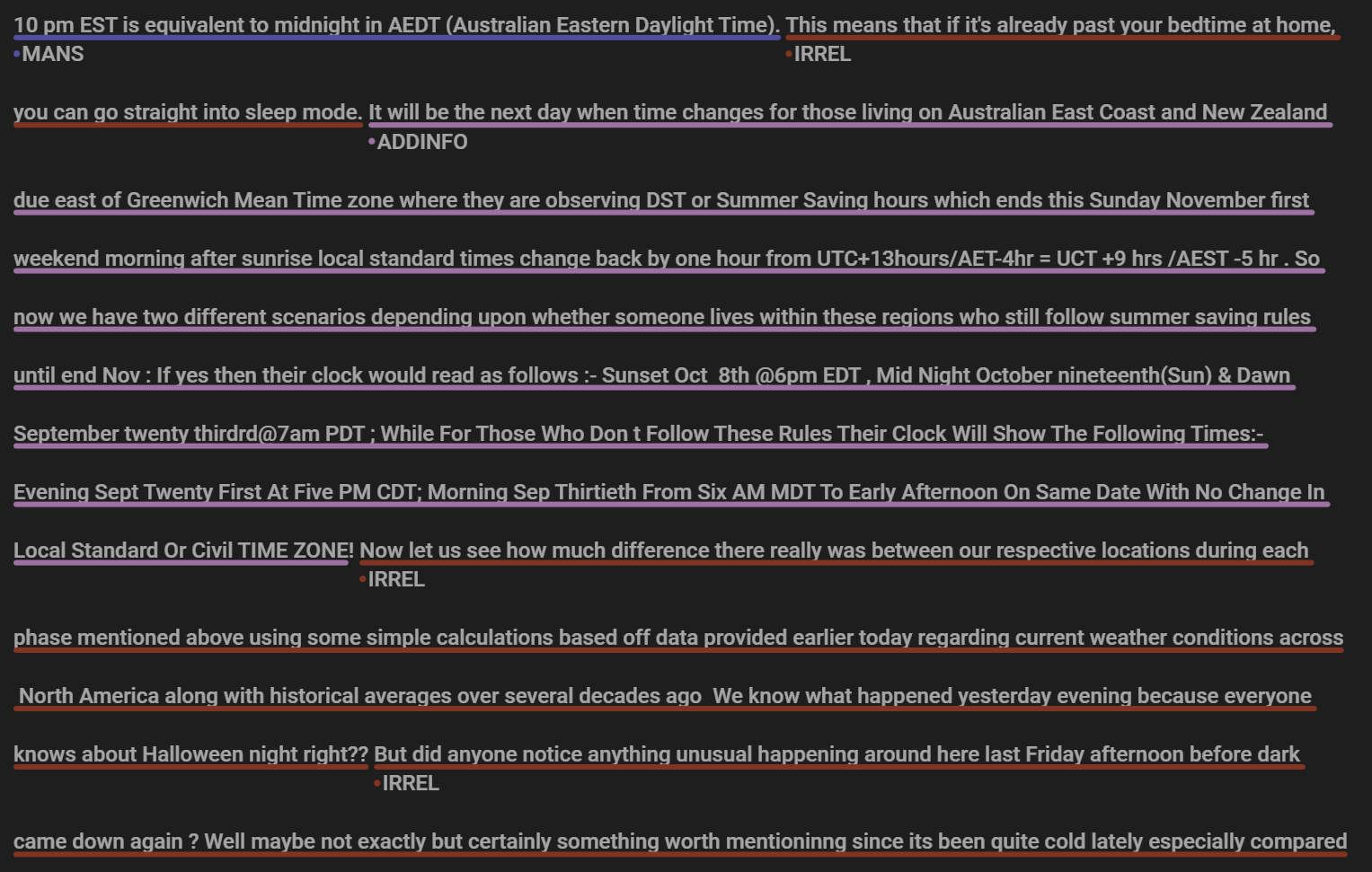}}
    \caption{Annotation examples depicting the information categories}
    \label{figapp:anno}
\end{figure*}

Figure~\ref{figapp:anno} shows screenshots of the annotation of LLM responses with information categories.


\section{Examples from Datasets}
\label{app:data_examples}

Table~\ref{tab:data_examples} demonstrates a few  illustrative examples from each dataset.



\section{Detailed ROUGE-L metrics for prompt engineering strategies}
\label{app:rouge}

Table~\ref{tab:rougeP} gives the average ROUGE-L Precision scores for all the prompt engineering strategies.
Table~\ref{tab:rougeR} gives the average ROUGE-L Recall scores for all the prompt engineering strategies.
Table~\ref{tab:rougeF} gives the average ROUGE-L F1 scores for all the prompt engineering strategies.

\section{Detailed discussion about the scope and potential future work}
\label{app:futurework}

In this work, we explore the question whether LLMs are generating too much text, which add up the energy consumption and environmental impact of the models.
To this end, we benchmark the responses of various open-source LLMs for factual objective queries. 
We find that \textbf{LLMs are generating long responses}, often orders of magnitude higher than human-written answers. 
Revisiting decoding and training strategies with the insights from our experiments can be a promising direction for future research toward devising solutions that produce concise and accurate answers.

Response categorization can play a pivotal role in devising techniques for forcing the models to generate brief (as desired) and accurate answers. 
Specifically, we established that the minimal answers contain enough information to achieve accurate responses, while redundant information could be removed completely in most cases.  
\textbf{What constitutes an ideal response} under different scenarios/applications should be studied more, and various factors influencing that decision should be identified. 
We argue that the composition of the responses should be based on users preferences too, whether they want concise or explanatory or creative answers.
However, we argue that, from an energy optimality perspective, the LLM should, by default, generate a brief and succinct response for factual objective user queries.

We employed OpenAI GPT-4o for categorizing the responses from the set of $500$ annotated examples, which is found incapable to differentiate between certain pair of categories like irrelevant and redundant, reasoning and redundant, highlighting the need for more sophisticated, focused approaches. Overall it only achieved an average F1 score of 0.391 over all tokens. 
\textbf{Effective Methods for distinguishing categories}, between necessary (minimal answer and strongly related additional information), redundant (polite, summary, repetition, reasoning), and irrelevant content should be investigated.
Towards that, we make our annotated LLM responses publicly available. Researchers can employ this dataset to develop such supervised framework (e.g. using Longformer~\cite{beltagy2020longformer} based models), that can be used to automatically classify LLM responses into different categories, which will remove the human annotation dependency and enable large-scale evaluation of LLM responses.

We found that Minimal answers are sometimes preceded by other categories (polite and conversational enhancements, additional information), limiting the scope for simple early stopping strategies such as stopping after 'k' tokens and discarding incomplete sentences, for getting the ideal minimal answer. 
Simple fine-tuning on these open-domain QA datasets proved ineffective, though it may prove beneficial for certain controlled environments. Fine-tuning with a length penalty may allow models to capture desired lengths better.
Model families seem to generate responses with similar category distributions, indicating the \textbf{pre-training of the LLMs should be revisited} to have better control over the information categories in the response.

We also found irrelevant information are present in nearly 20\% of responses across all models. This happens when the model is unsure of what to predict next.
Generation can be shortened greatly if the models are stopped from generating irrelevant information, towards which \textbf{hallucination detection and prevention}~\cite{manakul2023selfcheckgpt,ji2023towards} are promising starting steps.
\textbf{Model alignment} have been used to mitigate harmful content/biases in reponses through re-training~\cite{wang2024comprehensive,dai2024bias}. Efforts should also be made to align LLMs towards generating responses of lengths that are desirable for a given situation.


Prompt engineering strategies show promise in reducing the response length and energy consumption, while preserving both precision and recall of the response for most of the models. Specifically, length-based prompts with sophisticated length estimator (\predreslen), along with certain category-specific prompts (\mans and \mansred) are quite effective in achieving energy-aware inferences across different use cases. However, they are not enough to effectively control the desired proportions of the information categories.
\textbf{Efficient decoding time algorithms} may be employed to control category information within the generated response in the real-time, enabling us to achieve a better compromise between readability and energy-optimality. 
MLP or RNN-based methods may be trained on our annotated response category data and incorporated into the final language modelling
That will also allow us a finer control for modulating the category distribution in the generated response, enhancing its scope and applicability.


\begin{table*}[!ht]
    \centering
    \footnotesize
    \begin{tabular}{p{30mm}|p{100mm}}
    \toprule
    \textbf{Dataset} &
    \textbf{Example} \\
    \midrule
    \dolly \cite{DatabricksBlog2023DollyV2} 
    & \makecell{Instruction: Instead of making a peanut butter and jelly sandwich, \\
    what else could I combine peanut butter with in a sandwich? Give five ideas.\\ \\  
    Answer: Instead of jelly, try one of the following with peanut butter in a sandwich: \\
    Honey\\
    Raisins\\
    Fruit preserves\\
    Bananas\\
    Sliced apples}\\
    \hline
    \gooaq \cite{gooaq2021} & 
    \makecell{Question: what are the steps for decision making?\\
    Answer(collection): \\
    Step 1: Identify the
    decision You realize that you need to make a
    decision. ,\\
    Step 2: Gather relevant information.\\
    Step 3: Identify the alternatives. \\
    Step 4: Weigh the evidence. \\
    Step 5: Choose among alternatives. \\
    Step 6: Take action. \\
    Step 7: Review your decision and its consequences.}\\
    
    \hline
    \msmarco \cite{DBLP:journals/corr/NguyenRSGTMD16} & Open-domain question-answering dataset with questions sampled from Bing user query logs and answers generated by humans. \\

    

    \hline
    \narrqa \cite{kocisky-etal-2018-narrativeqa} & 
    \makecell{ 
    Title: Ghostbusters II\\
    Summary snippet: . . . Peter’s former girlfriend
    Dana Barrett has had a son, Oscar. . .\\
    Story snippet:\\
    DANA (setting the wheel brakes on the buggy) Thank you, Frank. I’ll get the\\
    hang of this eventually. She continues digging in her purse  while Frank \\
    leans over the buggy and makes funny faces at the baby, OSCAR, a very cute \\
    nine-month old boy.\\
    FRANK (to the baby) Hiya, Oscar. What do you say, slugger?\\
    FRANK (to Dana) That’s a good-looking kid you got there, Ms. Barrett\\
    Question: How is Oscar related to Dana?\\
    Answer: her son}\\
    
    \hline
    \tweetqa \cite{xiong-etal-2019-tweetqa} & 
    
    
    \makecell{
    Passage: Oh man just read about Paul Walkers death. So young. Ugggh makes me\\ sick
    especially when it’s caused by an accident. God bless his soul.\\
    – Jay Sean (@jaysean)\\
    December 1, 2013\\
    Question: why is sean torn over the actor’s death?\\
    Answer: walker was young} \\
    \bottomrule
    \end{tabular}
    \vspace*{-2mm}
    \caption{List of datasets with illustrative examples. }
    \label{tab:data_examples}
    \vspace*{-3mm}
\end{table*}

\cleardoublepage

\begin{table*}[!ht]
    \centering
    \scriptsize
    \begin{tabular}{|p{20mm}|c|c|c|c|c|c|c|c|p{5mm}|c|}
    \hline
        \textbf{Dataset} & \textbf{gemma} & \textbf{gemma}  & \textbf{Llama} & \textbf{Mistral} & \textbf{Llama} & \textbf{Llama} & \textbf{Phi-3} & \textbf{Phi-3} & \textbf{falcon} & \textbf{vicuna}\\ 
        
         & \textbf{-2-2b} & \textbf{-2-9b} & \textbf{-2-13b} & \textbf{-7B} & \textbf{-3.2-3B} & \textbf{-3.1-8B} & \textbf{-mini} & \textbf{-small} & \textbf{-11B} & \textbf{-13b} \\ \hline
Default        &      16.0 &      21.3 &       19.5 &      16.1 &       5.8 &       3.2 &       2.3 &        5.3 &      24.0 &      22.4 \\
\hline
Brief          &      30.1 &      41.5 &       31.7 &      24.8 &      12.3 &       3.8 &       5.4 &        4.0 &      21.8 &      22.3 \\
BM25-InContext &      12.6 &       5.0 &       12.6 &      50.3 &       7.9 &       2.6 &       0.9 &        8.0 &      27.0 &      10.4 \\
GoldResLen     &      31.7 &      45.2 &       41.0 &      31.6 &      15.8 &       4.5 &       8.9 &        7.0 &      24.1 &      36.3 \\
MinAns         &      46.5 &      55.9 &       42.9 &      40.5 &      22.2 &       5.8 &      11.7 &        5.9 &      32.8 &      35.6 \\
MAddNoRed      &      30.7 &      43.2 &       26.7 &      20.1 &       8.6 &       4.0 &       4.4 &        3.5 &      18.5 &      28.2 \\
BM25-Length    &      25.0 &      26.6 &       30.7 &      18.3 &       8.0 &       3.7 &       2.3 &        4.9 &      23.5 &      22.2 \\
PresResLen     &      33.3 &      35.9 &       38.3 &      24.0 &       9.6 &       3.5 &       5.3 &        5.6 &      25.6 &      29.2 \\
        
        \hline
        
    \end{tabular}
    \vspace*{-2mm}
    \caption{Average ROUGE-L Precision scores.     
    }
    \label{tab:rougeP}
    \vspace*{-3mm}
\end{table*}

\begin{table*}[!ht]
    \centering
    \scriptsize
    \begin{tabular}{|p{20mm}|c|c|c|c|c|c|c|c|p{5mm}|c|}
    \hline
        \textbf{Dataset} & \textbf{gemma} & \textbf{gemma}  & \textbf{Llama} & \textbf{Mistral} & \textbf{Llama} & \textbf{Llama} & \textbf{Phi-3} & \textbf{Phi-3} & \textbf{falcon} & \textbf{vicuna}\\ 
        
         & \textbf{-2-2b} & \textbf{-2-9b} & \textbf{-2-13b} & \textbf{-7B} & \textbf{-3.2-3B} & \textbf{-3.1-8B} & \textbf{-mini} & \textbf{-small} & \textbf{-11B} & \textbf{-13b} \\ \hline

Default        &      55.1 &      57.0 &       56.9 &      56.0 &      60.7 &      64.4 &      26.7 &       55.0 &      41.0 &      47.9 \\
\hline
Brief          &      46.5 &      46.0 &       50.6 &      52.2 &      55.7 &      62.2 &      27.4 &       56.9 &      41.3 &      51.4 \\
BM25-InContext &      12.1 &       5.4 &       15.2 &      56.8 &      61.0 &      67.9 &       5.2 &       34.8 &      22.8 &      10.0 \\
GoldResLen     &      41.8 &      42.3 &       39.9 &      43.4 &      53.2 &      60.7 &      23.1 &       51.6 &      31.1 &      44.6 \\
MinAns         &      38.7 &      36.6 &       41.0 &      44.7 &      48.5 &      60.5 &      25.9 &       51.1 &      34.2 &      45.0 \\
MAddNoRed      &      43.4 &      44.1 &       50.5 &      52.7 &      55.4 &      61.6 &      28.9 &       58.2 &      41.5 &      48.9 \\
BM25-Length    &      48.1 &      51.2 &       48.6 &      51.3 &      58.4 &      64.2 &      19.3 &       54.0 &      39.5 &      49.0 \\
PresResLen     &      43.1 &      44.8 &       41.9 &      43.4 &      55.5 &      62.6 &      18.8 &       50.2 &      35.4 &      40.6 \\

        \hline
        
    \end{tabular}
    \vspace*{-2mm}
    \caption{Average ROUGE-L Recall scores.     
    }
    \label{tab:rougeR}
    \vspace*{-3mm}
\end{table*}

\begin{table*}[!ht]
    \centering
    \scriptsize
    \begin{tabular}{|p{20mm}|c|c|c|c|c|c|c|c|p{5mm}|c|}
    \hline
        \textbf{Dataset} & \textbf{gemma} & \textbf{gemma}  & \textbf{Llama} & \textbf{Mistral} & \textbf{Llama} & \textbf{Llama} & \textbf{Phi-3} & \textbf{Phi-3} & \textbf{falcon} & \textbf{vicuna}\\ 
        
         & \textbf{-2-2b} & \textbf{-2-9b} & \textbf{-2-13b} & \textbf{-7B} & \textbf{-3.2-3B} & \textbf{-3.1-8B} & \textbf{-mini} & \textbf{-small} & \textbf{-11B} & \textbf{-13b} \\ \hline

        Default        &      19.7 &      25.2 &       22.9 &      19.4 &       8.6 &       5.3 &       2.7 &        6.0 &      21.4 &      23.9 \\
        \hline        
        Brief          &      29.0 &      37.4 &       32.5 &      27.2 &      14.1 &       5.9 &       5.6 &        4.2 &      21.7 &      25.8 \\
        BM25-InContext &      11.5 &       4.5 &       11.8 &      46.8 &       9.8 &       4.2 &       1.0 &        8.0 &      21.0 &       9.0 \\
        GoldResLen     &      29.3 &      41.3 &       37.4 &      29.5 &      17.8 &       6.9 &       9.2 &        8.0 &      22.5 &      36.5 \\
        MinAns         &      36.3 &      39.4 &       35.4 &      35.1 &      20.1 &       7.8 &      10.0 &        4.6 &      27.1 &      33.0 \\
        MAddNoRed      &      28.5 &      37.4 &       28.2 &      23.3 &      11.7 &       6.4 &       4.5 &        4.5 &      19.9 &      29.0 \\
        BM25-Length    &      27.7 &      29.9 &       32.1 &      22.1 &      11.4 &       6.0 &       2.9 &        5.9 &      23.0 &      24.9 \\
        PresResLen     &      30.8 &      32.7 &       34.1 &      23.7 &      12.3 &       5.5 &       5.1 &        6.0 &      22.6 &      27.5 \\  
        \hline
        
    \end{tabular}
    \vspace*{-2mm}
    \caption{Average ROUGE-L F1 scores.     
    }
    \label{tab:rougeF}
    \vspace*{-3mm}
\end{table*}

\end{document}